%% file: paper.tex
\documentclass[runningheads,a4paper]{llncs}
\usepackage{graphicx}
\usepackage{times}
\usepackage{epsfig}
\usepackage{amsmath}
\usepackage{amssymb}
\usepackage{array}
\usepackage{booktabs}
\usepackage{cite}

\usepackage{floatrow}
\newfloatcommand{capbtabbox}{table}[][0.5\textwidth]

\usepackage{url}
\urldef{\mailsa}\path|ignaciorlando@gmail.com|
\urldef{\mailsb}\path||
\urldef{\mailsc}\path||    
\newcommand{\keywords}[1]{\par\addvspace\baselineskip
\noindent\keywordname\enspace\ignorespaces#1}

\begin{document}

\mainmatter  

\title{Arabidopsis roots segmentation based on morphological operations and CRFs}
\titlerunning{Arabidopsis roots segmentation based on morphological operations and CRFs}

%
%
\author{Jos\'e Ignacio Orlando\inst{1,2}, Hugo Luis Manterola\inst{1,2}, Enzo Ferrante\inst{3} and~Federico~Ariel\inst{4,5}}
\authorrunning{Jos\'e Ignacio Orlando, Hugo Luis Manterola, Enzo Ferrante and Federico Ariel}

\institute{{$^{1}$ Pladema Institute, UNCPBA, Argentina}\\
{$^{2}$ Consejo Nacional de Investigaciones Cient\'ificas y T\'ecnicas, CONICET, Argentina}\\
{$^{3}$ Center for Visual Computing, \'Ecole Centrale Paris, France}\\
{$^{4}$ CNRS, Institut des Sciences du V\'eg\'etal, Saclay Plant Sciences, Gif sur Yvette, France}\\
{$^{5}$ Universit\'e Paris Diderot, Paris, France}\\
}

%
%

\toctitle{Arabidopsis roots segmentation based on morphological operations and CRFs}
\tocauthor{Jos\'e Ignacio Orlando, Hugo Luis Manterola, Enzo Ferrante and Federico Ariel}
\maketitle
\input{abstract.tex}
\input{introduction.tex}

\input{preprocessing.tex}
\input{segmentation.tex}
\input{postprocessing.tex}
\input{results.tex}

\input{conclusions.tex}

\bibliography{rootSegmentation}

\end{document}

%% file: abstract.tex
\begin{abstract}

\textit{Arabidopsis thaliana} is a plant species widely utilized by scientists to estimate the impact of genetic differences in root morphological features. For this purpose, images of this plant after genetic modifications are taken to study differences in the root architecture. This task requires manual segmentations of radicular structures, although this is a particularly tedious and time-consuming labor. In this work, we present an unsupervised method for \textit{Arabidopsis thaliana} root segmentation based on morphological operations and fully-connected Conditional Random Fields. Although other approaches have been proposed to this purpose, all of them are based on more complex and expensive imaging modalities. Our results prove that our method can be easily applied over images taken using conventional scanners, with a minor user intervention. A first data set, our results and a fully open source implementation are available online.

\keywords{Roots segmentation, Automatic phenotyping, Morphological operations, Conditional Random Fields.}
\end{abstract}

%% file: introduction.tex
\section{Introduction}
\label{sec:introduction}

\textit{Arabidopsis thaliana} is a small flowering plant species native to Eurasia, and up to date it is the most popular model organism in plant molecular biology and genetics \cite{arabidopsis2000analysis}. This plant was the first to have its genome sequenced, and is a unique tool for understanding the molecular
biology of several plant traits, including the analysis of root behaviour under a wide variety of environmental cues \cite{granier2006phenopsis}.

Analysing root system architecture of plants is of key interest in several scientific areas, including molecular biology, plant physiology and biotechnology \cite{pound2013rootnav}. First, because it allows to estimate the impact of genetic differences in root phenotype (e. g., in the analysis of mutant and transgenic plants, or ecotypes of the same species in similar environmental conditions). On the other hand, it also enables scientists to quantify how environmental conditions-–such as drought or toxic agents in the soil–-affect plants with the same genetic background \cite{fitter1986topology}.

One of the simplest ways to quantify a morphological phenotype of the root architecture consists in analysing root photographs \cite{lobet2011novel}. Counting on precise segmentations of these roots is crucial to extract features such as length of the main root, number of branches (so-called lateral roots), angle of curvature, among others. Those features served to describe in detail the roots morphology, and can be associated to genetic changes and/or in response to environmental conditions \cite{pound2013rootnav}. Up to now, current best practice consists first in segmenting the root architecture manually, although it is a tedious and time consuming task. After segmentation, features are also extracted manually, and the corresponding intra- and inter- observer differences may result in a remarkable variation, which can conduce to errors in the subsequent experimental analysis \cite{clark2013high}. 

In this work, we focus on the task of segmenting, with a minor user intervention,
the root architecture of \textit{Arabidopsis thaliana}. Alternative tools are available nowadays
to segment and analyze root architectures. GiA Roots \cite{gal12} and RootReader2D \cite{cla13} use various segmentation methods based on thresholding. The main problem with
this simple approach is that it requires high contrast between the roots and their background to
achieve good results. Rootnav \cite{pou13} uses a more sophisticated method based on the expectation maximization classification algorithm to examine the input image, obtaining the
likelihood of each pixel belonging to a root. This tool allows the user to apply corrections
over the estimated root model to avoid potential mistakes. However, several details still remain to be improved, such as the case of crossed lateral roots and the corresponding main root association. Other approaches as Level Sets \cite{mai12} and livewire algorithm \cite{bas12} are also proposed in the literature as alternative approaches to this task.

Conditional Random Fields (CRFs) are widely utilized in several computer vision and image processing applications \cite{li2009markov}. In the last few years, Kr\"ahenb\"uhl and Koltun have introduced a novel technique for efficient inference of fully- connected CRFs with Gaussian edge potentials \cite{krahenbuhl2012efficient}. The low computational cost of that novel approach have been recently taken into account by Orlando and Blaschko \cite{orlando2014learning}, who use a Structured Output SVM to train fully-connected CRFs for blood vessel segmentation in retinal images. Authors have claimed that the rich information provided by the combination of
their discriminatively training method and the fully-connectedness of the pairwise potentials can be exploited for automatic detection of elongated structures \cite{orlando2014learning}. However, supervised methods cannot be applied in our case due to the small size of our data set.

In this paper we propose a first evaluation of an unsupervised approach for \textit{Arabidopsis thaliana} root segmentation based on morphological operations and fully- connected CRFs. Unlike other approaches based on complex and expensive imaging modalities such as nuclear magnetic resonance, X-ray computed tomography or laser scanning \cite{galkovskyi2012gia}, our method 
 is applied over photographs obtained by scanning the plates of seedlings. Our main goal is to develop a free open source tool for automatic root phenotyping that allows biologists to obtain precise information about root architecture without any intervention.

\begin{figure}[t]
  \centering
    \includegraphics[scale=0.70]{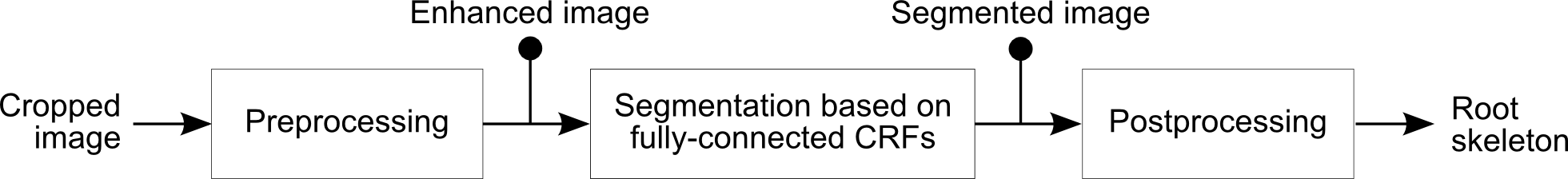}
  \caption{Proposed segmentation approach.}
	\label{fig:pipeline}
\end{figure}

A general view of our method is summarized in Fig. \ref{fig:pipeline}. Given a photograph of a plant cropped by the user, we first preprocessed it to remove the leaves and to enhance the connectivity between different parts of the root architecture. Using the resulting image to generate unary and pairwise potentials on a fully-connected CRF we obtain an initial segmentation, which is then postprocessed to improve connectivity between isolated segments. Finally, a skeletonization algorithm is applied to get root centerlines, and small segments are cleaned to get the final segmentation. We find that this first simple approach provides promising results when applied over real images, with a low computational cost. We also release our data set in order to encourage other scientists to evaluate their own algorithms.

The remainder of this paper is organized as follows: Section~\ref{sec:preprocessing} refers to the preprocessing operations applied over images to facilitate the posterior segmentation task. The formulation of fully-connected CRFs is analysed in detail in Section~\ref{sec:segmentation}. Section~\ref{sec:postprocessing} depicts methods applied to postprocess the resulting segmentation. In Section~\ref{sec:experiments} we describe the dataset we utilize to evaluate our method, and we include our results. Finally, Section~\ref{sec:conclusions} concludes the paper.

%% file: preprocessing.tex
\section{Preprocessing}
\label{sec:preprocessing}

\begin{figure}[t]
  \centering
    \includegraphics[scale=0.30]{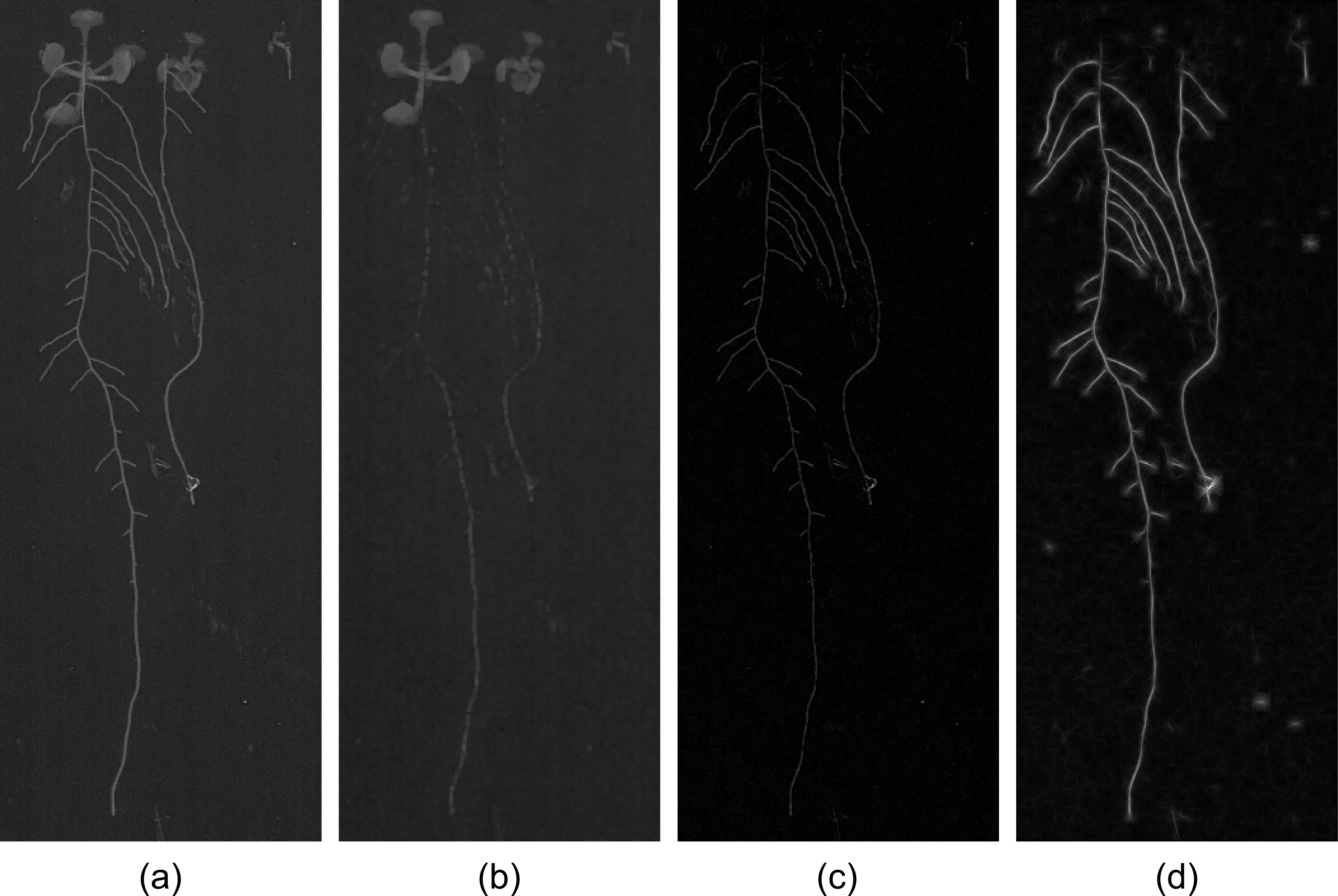}
  \caption{Preprocessing of a root image. (a) Original image. (b) Leaves obtained by erosion and dilation. (c) Leaves subtracted image. (d) Enhanced image after applying multiple line detectors.}
	\label{fig:preprocessedRoot}
\end{figure}

Original photographs usually contain more than just a single plant. Thus, it is necessary to crop the image manually to get a first approximation to the roots to be analyzed. Afterwards, intensities are mapped to the interval $[0,255]$ to increase the contrast between roots and background. This operation is performed by doing:
\begin{equation}
I_c = 255 \left( \frac{I - \min\{I\}}{\max\{I\} - \min\{I\}} \right)
\end{equation}
where $I$ is the original cropped image and $I_c$ is the contrast-enhanced image. An example of the resulting image can be observed in Fig. \ref{fig:preprocessedRoot} (a).  

Leaves appear at the top of the image as ellipsoidal blobs and have similar intensities than the roots. This property can produce false positives in the final segmentation. To reduce their influence we remove the leaves using morphological operations. Firstly, erosion with square structuring elements of side 3 is applied to remove roots. The resulting image is then dilated using the same structuring element, recovering the leaves as in Fig. \ref{fig:preprocessedRoot} (b). Finally, this image is subtracted to $I_c$, resulting in an image as the one in Fig. \ref{fig:preprocessedRoot} (c).

The leaves subtraction operation not only eliminates the leaves but also reduces the contrast between roots and background. By applying a variation of the line detectors introduced by Ricci and Perfetti in \cite{ricci2007retinal} for blood vessels segmentation in retinal images, roots contrast is enhanced. Authors propose to evaluate the gray level along lines of fixed length $l$ passing through each target pixel $(i,j)$ in the image at different orientations. In particular, angles from 0 to 165 degrees with an angular resolution of 15 degrees are employed in this work. The line of length $l$ with the largest average gray level $L_l(i,j)$ is then found. The line strength of the pixel is obtained as the difference $S_l(i,j) = L_l(i,j) - N_l(i,j)$, where $N_l(i,j)$ represents the average gray level in a square window of side $l$. Line strengths $S_l$ are obtained for each pixel in the image at different length resolutions, moving from $l=3$ to 15 with a step of 2. The maximum over all resolutions is taken as the new enhanced image $I_H$:
\begin{equation}
I_H(i,j) = \max_l \{ S_l(i,j) \}
\end{equation}
An example of $I_H$ can be observed in Fig. \ref{fig:preprocessedRoot} (d).

%% file: segmentation.tex
\section{Root segmentation based on fully-connected CRFs}
\label{sec:segmentation}

Many computer vision works pose the segmentation process as an energy minimization problem in a fully-connected Conditional Random Field (CRF). In this type of models, the image is mapped to a graph where each pixel represents a node and each node is connected to its neighbours by an edge. A fully-connected version assumes that every pixel is connected to every other through an edge. As it was shown by Kr\"ahenb\"uhl and Koltun in \cite{krahenbuhl2012efficient}, this property allows to take into account not only the local information provided by the interactions between neighborhoods in the image but also about long-range interactions, which proves to be effective when dealing with elongated structures \cite{orlando2014learning,orlando2017discriminatively}. In this work we follow the efficient inference approach presented in \cite{krahenbuhl2012efficient}, based on a mean field approximation of the original CRF.

Given an image labeling $\mathbf{y}$, its energy is obtained as the sum of its unary potentials $\psi_u$ and its pairwise potentials $\psi_p$:
\begin{equation}
E(\mathbf{y}) = \sum_i \psi_u(x_i, y_i) + \sum_{i<j} \psi_p(x_i,x_j,y_i, y_j)
\end{equation}
where $x_i$ is the value of a feature on the $i$-th pixel. In our case, we utilize the enhanced image $I_H$ as the feature for both potentials.

Unary potentials define a log-likelihood over the labeling $\mathbf{y}$, and are usually determined by a classifier \cite{krahenbuhl2012efficient}. In this work, unary potentials are given by the expression:
\begin{equation}
\psi_u(x_i,y_i) = \left\{ \begin{array}{lcc}
             w_u x_i &   \text{if}  & y_i=1 \\
             w_u \left( \max\{x\} - x_i \right) & \text{if} & y_i=-1 \\
             \end{array}
   \right.
\end{equation}
where $w_u$ is a weight parameter. 

Pairwise potentials define a similar distribution but considering interactions between pixels features and labels. In this paper we define our pairwise energy in a similar way than in \cite{orlando2014learning}:
\begin{equation}
\psi_p(x_i,x_j,y_i, y_j) = w_p \mu(y_i,y_j) \exp \left( -\frac{{|\mathbf{p}_i - \mathbf{p}_j|}^2}{2 \theta^2_p} - \frac{{|x_i - x_j|}^2}{2 \theta^2_x} \right)
\end{equation} 
where $w_p$ is a weight parameter, $\mu$ is a compatibility function and $\mathbf{p}_i$ and $\mathbf{p}_j$ are vectors with the spatial coordinates of pixels $i$ and $j$ respectively. The first term in the exponential controls the proximity of pixels: as the distance between pixels increases so does the difference between vectors $\mathbf{p}_i$ and $\mathbf{p}_j$ , and the energy $E$; analogously, when pixels are closer, the difference is smaller and the energy decreases. In a similar way, the second term refers to the similarity between pixel features. Parameters $\theta_p$ and $\theta_x$ control the degree of relevance of the proximity and the similarity parts, respectively. Finally, the compatibility function $\mu$ is given by the following expression:
\begin{equation}
\mu(y_i,y_j) = [y_i \neq y_j]
\end{equation}
meaning that it is 1 when labels are different and 0 otherwise. This property allows to penalize neighbours with different labels. 

\begin{figure}[t]
  \centering
    \includegraphics[scale=0.30]{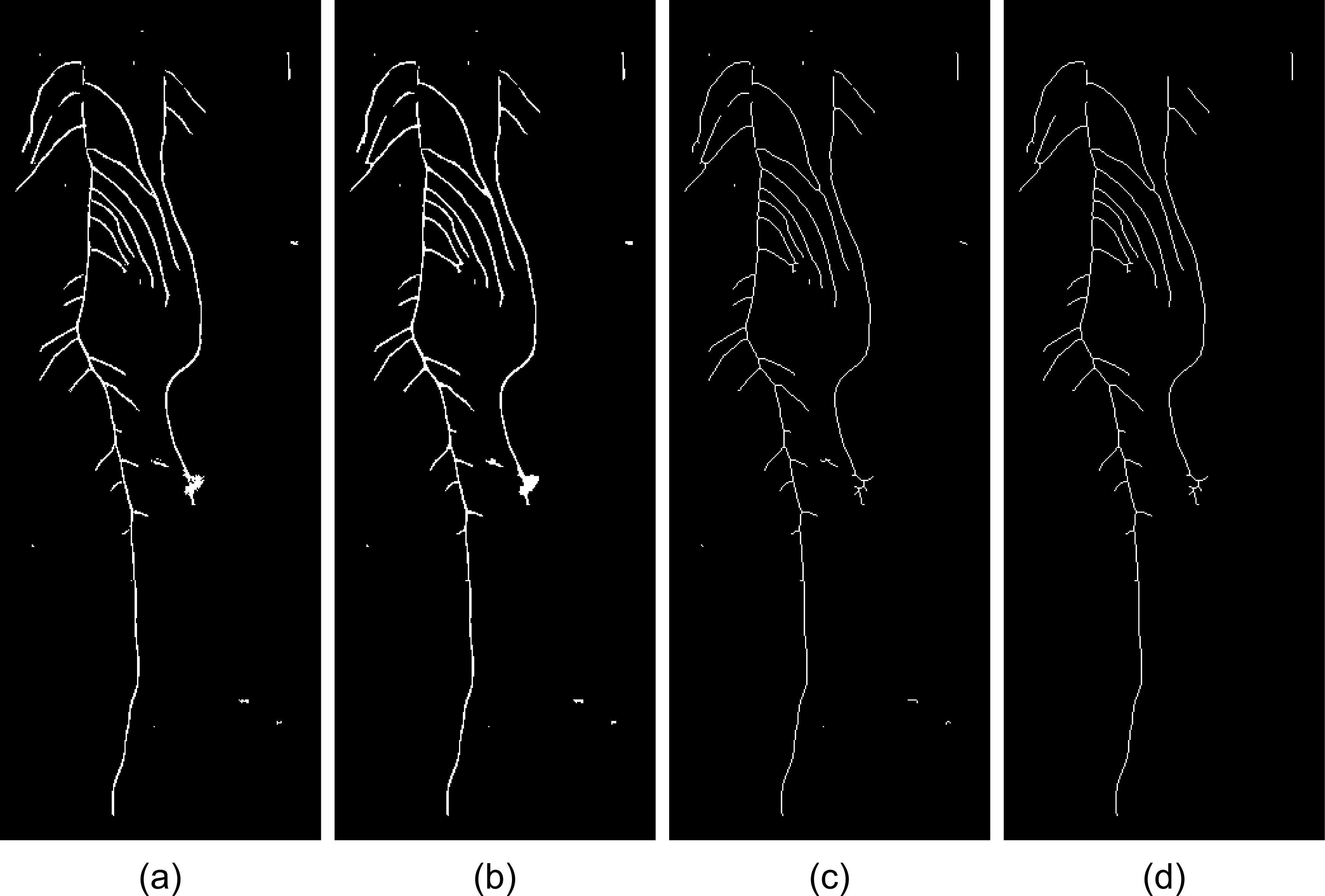}
  \caption{Segmentation and postprocessing of the segmented root. (a) Segmentation obtained using fully-connected CRFs. (b) Segmentation after filling gaps between pixels. (c) Skeletonization. (d) Skeletonization after removing short segments.}
	\label{fig:postprocessing}
\end{figure}

A segmentation example obtained using the fully-connected CRF can be observed in Fig. \ref{fig:postprocessing} (a).

%% file: postprocessing.tex
\section{Postprocessing}
\label{sec:postprocessing}

Segmentation obtained with the fully-connected CRF sometimes presents empty gaps between pixels corresponding to the same root. Those gaps are filled applying morphological closing with square structuring elements of side 3, resulting in masks as the one in Fig. \ref{fig:postprocessing} (b). Afterwards, the root skeleton is extracted considering 8-connectivity neighborhoods (Fig. \ref{fig:postprocessing} (c)). The final segmentation is obtained after filtering the skeleton according to an area criterion, based on removing those segments with an area lower than a given parameter $\alpha$. The area of a segment is evaluated as the sum of pixels of the segment in an 8-connectivity neighbourhood. An example of a final result is presented in Fig. \ref{fig:postprocessing} (d).

%% file: results.tex
\section{Validation and Results}
\label{sec:experiments}

We validate our method using a data set of 14 plant photographs. Each image contains 3 plants of different genetic backgrounds: \textit{Arabidopsis thaliana} ecotype ColO wild type, one RNAi line affecting the expression of the long non-coding RNA \textit{APOLO} \cite{ariel2014noncoding} and one line exhibiting an extreme phenotype, \textit{solitary root} (\textit{slr}, \cite{fukaki2002lateral}), which is impared in lateral root formation. Seedlings were grown in plates using the appropiate medium \cite{ariel2014noncoding}. 9 plants are included in each photograph. All images are in JPEG format, with 200 dpi resolution, and they were acquired using a conventional scanner. Manual annotations by two different human observers are available for quantitative evaluation. First observer segmentations are assumed as the ground truth, and the remaining are utilized as a reference of human observer performance. To evaluate skeletons with respect to the segmentations, both annotations were roughly delineated. The complete data set is available to be downloaded at \url{http://cvn.ecp.fr/personnel/eferrante/?page_id=534}.

We performed the evaluation of our method in terms of the following quality measure:
\begin{equation}
Q = \frac{| S \cap R |}{| S |}
\end{equation}
where $S$ is the final skeleton obtained using our approach and $R$ is the reference contour. This metric is an indicator of the overlap between the results and the gold-standard segmentation. 

\begin{table}
\begin{tabular}{| l || c |  c |  c || c |}
  \hline                       
  \textbf{Method}    & \textbf{Col0}     & \textbf{APOLO} & \textbf{slr}      & \textbf{Avg} \\
	\hline
	Our method         & 0.7425            & 0.8147              & 0.5872            & 0.7148$\pm$0.1627 \\
	\hline
	2nd human observer & 0.9225            & 0.9187              & 0.9243            & 0.9218$\pm$0.0367 \\
	\hline
\end{tabular}
	\caption{Average quality values of our method and the 2nd human observer.}
	\label{tab:results}
\end{table}

Average quality values obtained for each species of \textit{Arabidopsis thaliana} are listed in Table \ref{tab:results}. We also include the performance of the 2nd human observer for comparison purposes. Following \cite{krahenbuhl2012efficient} $\theta_p$ was set to 1. In the case of $\theta_x$, we follow the estimation approach proposed by Orlando and Blaschko in \cite{orlando2014learning}, based on taking the absolute value of the median of a random sample of 10.000 pairwise distances. As it was explained in Section~\ref{sec:segmentation}, parameters $w_u$ and $w_p$ weight unary and pairwise potentials, respectively, and determine their relevance in the energy function. Both parameters and the value of $\alpha$ were selected using the first image, so it was excluded from the evaluation. The best configuration according to our experiments is $w_u=2$, $w_p=1$ and $\alpha=20$.

%% file: conclusions.tex
\section{Conclusions}
\label{sec:conclusions}
In this work, we have presented a first evaluation of an unsupervised approach for \textit{Arabidopsis thaliana} root segmentation based on morphological operations and fully-connected CRFs. Our method is applied over photographs obtained by scanning the plates of seedlings. Results demonstrate that this approach could be an excellent starting point on developing a tool for automatic root phenotyping.  Features as length of the main root, number of lateral roots and angle of curvature can be automatically obtained from our segmentations.

Seedlings exhibiting extreme phenotypes regarding the average, likely due to seeds quality, are usually not considered for further analysis. However, we included those biased roots in the evaluation of our method. Future automatised image analysis procedures could potentially segment all roots and include all measurements of root morphological features, and discard those that are statistically out of the acceptable range.

Further parameters shall be included among those calculated in an automatised manner in the future, such as root architecture surface and geometry, distance between lateral root initiation points and root gravitropic response by angle calculation. Thus, informatic tools will enable plant scientists to describe the morphological phenotype of \textit{Arabidopsis} roots in a more comprehensive manner.\\~

\noindent
\textbf{Acknowledgements}\\
We thank Philip Kr\"ahenb\"uhl and Vladen Koltun for providing us with their code for efficient inference of fully-connected CRFs.